\documentclass[journal,twoside,web]{ieeecolor}
\usepackage{generic}
\usepackage{amsmath,amssymb,amsfonts}
\usepackage{algorithmic}
\usepackage{graphicx}
\usepackage{textcomp}

\usepackage{url}
\usepackage[hidelinks]{hyperref}
\usepackage{placeins}
\usepackage[T1]{fontenc}
\usepackage{hyperxmp}
\usepackage{booktabs}
\usepackage{balance}
 \usepackage{array,multirow}
 \usepackage{subcaption}

\newcommand{\sig}{\textsuperscript{$\dagger$}\hspace{0.0ex}}

\def\BibTeX{{\rm B\kern-.05em{\sc i\kern-.025em b}\kern-.08em
    T\kern-.1667em\lower.7ex\hbox{E}\kern-.125emX}}
\begin{document}
\title{Clinical Predictive Models for COVID-19: Systematic Study}

\author{Patrick Schwab, August DuMont Sch\"utte, Benedikt Dietz, and Stefan Bauer
\thanks{P. Schwab is with F. Hoffmann-La Roche Ltd, Basel, Switzerland, A. DuMont Sch\"utte and B. Dietz are with ETH Zurich, Switzerland, and S. Bauer is with the Max Planck Institute of Intelligent Systems, T\"ubingen, Germany (correspondence to: patrick.schwab@roche.com). }
}

\maketitle

\begin{abstract}
Coronavirus Disease 2019 (COVID-19) is a rapidly emerging respiratory disease caused by the severe acute respiratory syndrome coronavirus 2 (SARS-CoV-2). Due to the rapid human-to-human transmission of SARS-CoV-2, many healthcare systems are at risk of exceeding their healthcare capacities, in particular in terms of SARS-CoV-2 tests, hospital and intensive care unit (ICU) beds and mechanical ventilators. Predictive algorithms could potentially ease the strain on healthcare systems by identifying those who are most likely to receive a positive SARS-CoV-2 test, be hospitalised or admitted to the ICU. Here, we study clinical predictive models that estimate, using machine learning and based on routinely collected clinical data, which patients are likely to receive a positive SARS-CoV-2 test, require hospitalisation or intensive care. To evaluate the predictive performance of our models, we perform a retrospective evaluation on clinical and blood analysis data from a cohort of 5644 patients. Our experimental results indicate that our predictive models identify (i) patients that test positive for SARS-CoV-2 a priori at a sensitivity of 75\% (95\% CI: 67\%, 81\%) and a specificity of 49\% (95\% CI: 46\%, 51\%),  (ii) SARS-CoV-2 positive patients that require hospitalisation with 0.92 AUC (95\% CI: 0.81, 0.98), and (iii) SARS-CoV-2 positive patients that require critical care with 0.98 AUC (95\% CI: 0.95, 1.00). In addition, we determine which clinical features are predictive to what degree for each of the aforementioned clinical tasks. Our results indicate that predictive models trained on routinely collected clinical data could be used to predict clinical pathways for COVID-19, and therefore help inform care and prioritise resources.
\end{abstract} 


\section{Introduction}
\label{sec:introduction}
\IEEEPARstart{C}{oronavirus} Disease 2019 (COVID-19) was first discovered in December 2019 in China, and has since rapidly spread to over 200 countries \cite{who2020covid}. The COVID-19 pandemic challenges healthcare systems worldwide as a high peak capacity for testing and hospitalisation is necessary to diagnose and treat affected patients, particularly if the spread of SARS-CoV-2 is not mitigated. To avoid exceeding the available healthcare capacities, many countries have adopted social distancing policies, imposed travel restrictions, and postponed non-essential care and surgeries in order to reduce peak demand on their healthcare systems \cite{chinazzi2020effect,jernigan2020update,lin2020conceptual}.

\begin{figure}[t]
\centerline{\includegraphics[width=0.67\linewidth]{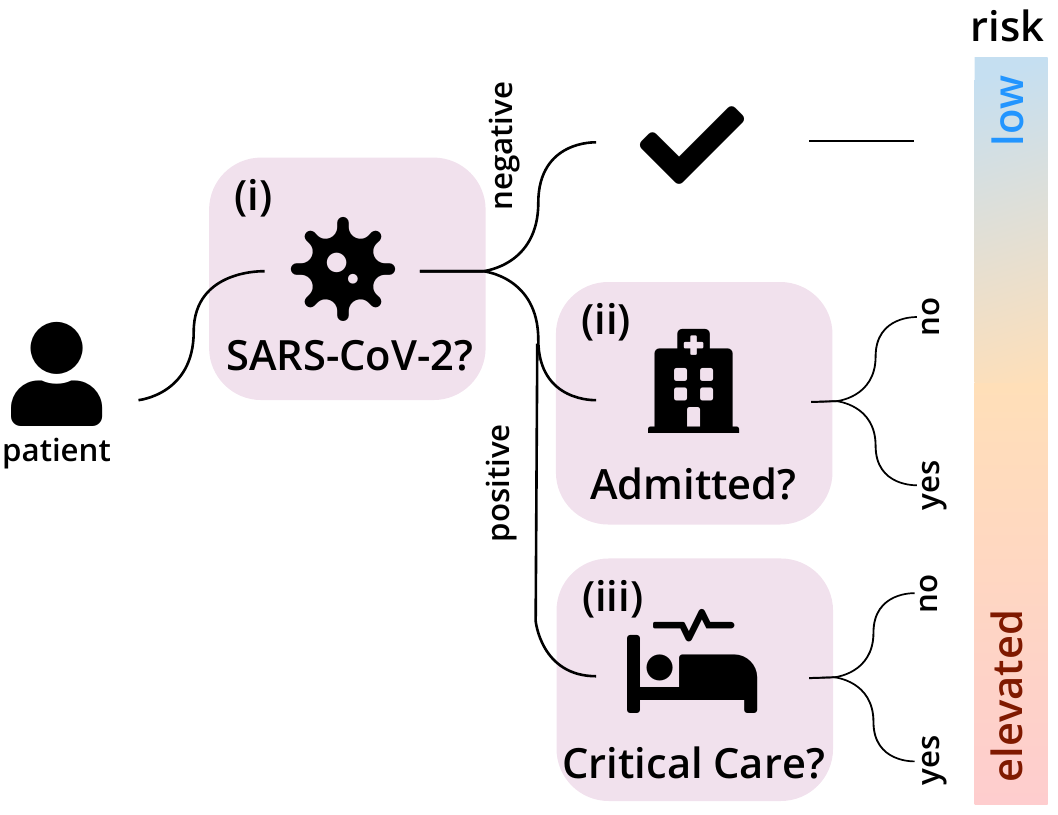}}
\caption{We study the use of predictive models (light purple) to estimate whether patients are likely (i) to be SARS-CoV-2 positive, and whether SARS-CoV-2 positive patients are likely (ii) to be admitted to the hospital and (iii) to require critical care based on clinical, demographic and blood analysis data. Accurate clinical predictive models stratify patients according to individual risk, and, in this manner, help prioritise healthcare resources, such as testing, hospital and critical care capacity.} 
\label{fig:teaser}
\end{figure}

The adoption of clinical predictive models that accurately predict who is likely to require testing, hospitalisation and intensive care  from routinely collected clinical data could potentially further reduce peak demand by ensuring resources are prioritised to those individuals with the highest risk (Figure \ref{fig:teaser}). For example, a clinical predictive model that accurately identifies patients that are likely to test positive for SARS-CoV-2 a priori could help prioritise limited SARS-CoV-2 testing capacity. However, developing accurate clinical prediction models for SARS-CoV-2 is difficult as relationships between clinical data, hospitalisation, and intensive care unit (ICU) admission have not yet been established conclusively due to the recent emergence of SARS-CoV-2.

In this systematic study, we develop and evaluate clinical predictive models that use routinely collected clinical data to identify (i) patients that are likely to receive a positive SARS-CoV-2 test, (ii) SARS-CoV-2 positive patients that are likely to require hospitalisation, and (iii) SARS-CoV-2 positive patients that are likely to require intensive care. Using the developed predictive models, we additionally determine which clinical features are most predictive for each of the aforementioned clinical tasks. Our results indicate that predictive models could be used to predict clinical pathways for COVID-19 patients. Such predictive models may be of significant utility for healthcare systems as preserving healthcare capacity has been linked to successfully combating SARS-CoV-2 \cite{wang2020response,lee2020interrupting}.

This work contains the following contributions:
\begin{itemize}
\item We develop and systematically study predictive models for estimating the likelihoods of (i) a positive SARS-CoV-2 test in patients presenting at hospitals, (ii) hospital admission in SARS-CoV-2 positive patients, and (iii) critical care admission in SARS-CoV-2 positive patients.
\item We validate the performance of the developed clinical predictive models in a retrospective evaluation using real-world data from a cohort of 5644 patients.
\item We determine and quantify the predictive power of routinely-collected clinical, demographic, and blood analysis data for the aforementioned clinical prediction tasks.
\end{itemize}

\section{Related Work}
A substantial body of work is dedicated to the study, validation and implementation of predictive models for clinical tasks. Clinical predictive models have, for example, been used to predict risk of septic shock \cite{henry2015targeted,horng2017creating}, risk of heart failure \cite{wu2010prediction}, readmission following heart failure \cite{frizzell2017prediction,shameer2017predictive,golas2018machine}, false alarms in critical care \cite{schwab2018not}, risk scores \cite{caruana2015intelligible}, outcomes \cite{visweswaran2005patient} and mortality in pneumonia \cite{cooper1997evaluation,wu2014using}, and mortality risk in critical care \cite{clermont2001predicting,ghassemi2014unfolding,johnson2017reproducibility}. Predicting clinical outcomes for individual patients is difficult because a large number of confounding factors may influence patient outcomes, and collecting and accounting for these factors in an unbiased way remains an open challenge in clinical practice \cite{obermeyer2016predicting}. Systematic studies, such as the one presented in this work, enable medical practitioners to better understand, assess and potentially overcome these issues by systematically evaluating generalisation ability, expected predictive performance, and influential predictors of various clinical predictive models. Beyond the need for systematic evaluation, missingness \cite{faris2002multiple,wells2013strategies,lipton2016directly,che2018recurrent}, noise \cite{lasko2013computational,schwab2017beat}, multivariate input data \cite{schwab2018not,schwab2019phonemd,ghassemi2015multivariate,schwab2020deepms}, and the need for interpretability \cite{choi2016retain,doshi2017towards,ross2017right,schwab2019cxplain} have been highlighted as particularly important considerations in healthcare settings. In this work, we build on recent methodological advances to develop and systematically study clinical predictive models that may aid in prioritising healthcare resources \cite{chen2017machine} for COVID-19, and thereby help prevent a potential overextension of healthcare system capacity. 

\subsection{Clinical Predictive Models for COVID-19} Several clinical predictive models have recently been proposed for COVID-19, for example, for predicting potential COVID-19 diagnoses using data from emergency care admission exams \cite{de2020covid} and chest imaging data \cite{wang2020covid,narin2020automatic,li2020artificial,castiglioni2020artificial,wang2020deep,xu2020deep}, for predicting COVID-19 related mortality from clinical risk factors \cite{sarkar2020machine,yan2020prediction}, and for predicting which patients will develop acute respiratory distress syndrome (ARDS) from patients' clinical characteristics \cite{jiang2020towards}. \cite{siordia2020epidemiology} presented a review of epidemiology and clinical features associated with COVID-19, and \cite{wynants2020prediction} a critical review that assessed limitations and risk of bias in diagnostic and prognostic models for COVID-19. In addition, \cite{wang2020clinical} performed a cohort study for clinical and laboratory predictors of COVID-19 related in-hospital mortality that identified baseline neutrophil count, age and several other clinical features as top predictors of mortality. Beyond prediction, \cite{ienca2020responsible} have argued for the responsible use of data in tackling the challenges posed by SARS-CoV-2.

Owing to the recent emergence of SARS-CoV-2, there currently exists, to the best of our knowledge, no prior systematic study on clinical predictive models that predict likelihood of a positive SARS-CoV-2 test, hospital and intensive care unit admission from clinical, demographic and blood analysis data that accounts for the missingness that is characteristic for the clinical setting. We additionally assess the influence of various clinical, demographic, and blood analysis measurements on the predictions of the developed clinical predictive models. 

\begin{figure}[t!]
\centerline{\includegraphics[width=\linewidth]{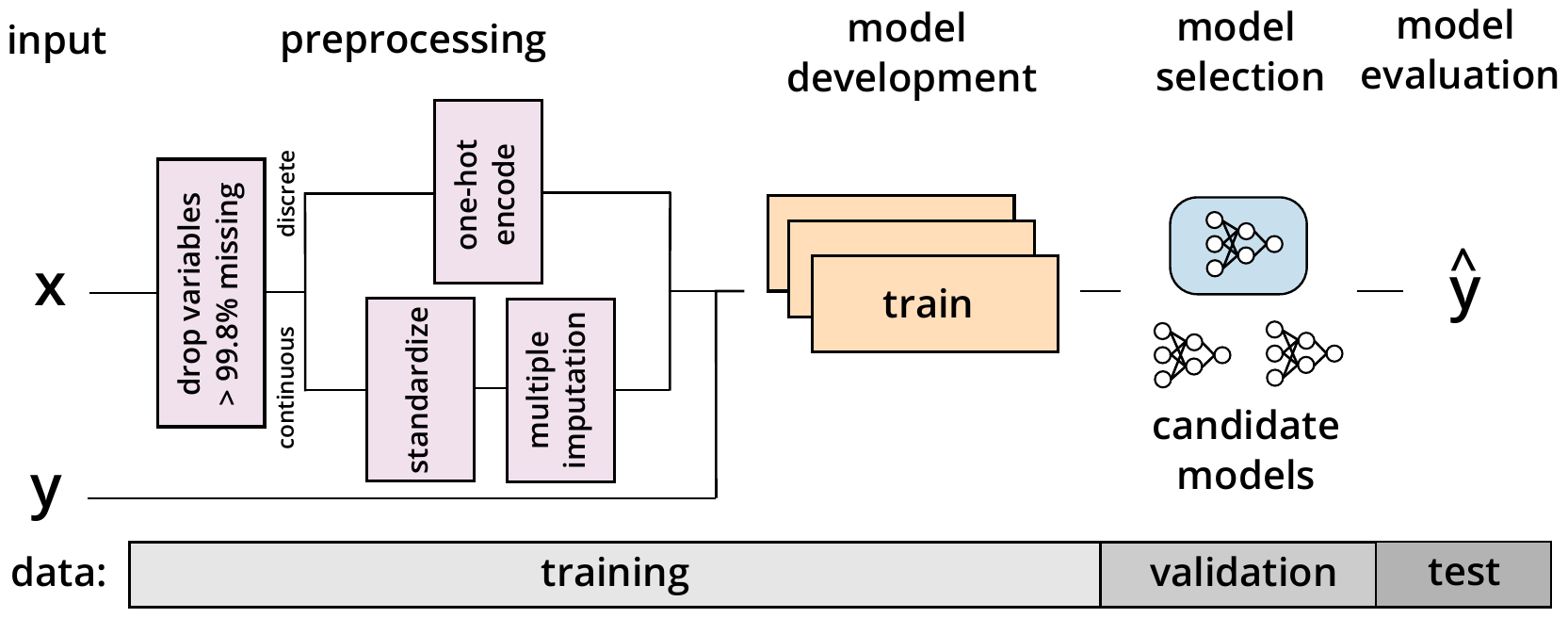}}
\caption{The presented multistage machine-learning pipeline consists of preprocessing (light purple) the input data $x$, developing multiple candidate models using the given dataset (orange), selecting the best candidate model for evaluation (blue), and evaluating the selected best model's outputs $\hat{y}$.} 
\label{fig:pipeline}
\end{figure}

\section{Methods and Materials}
\subsubsection{Problem Setting} In the given setting, we are given 106 routine clinical, laboratory and demographic measurements, or features, $x_i \in x$ for presenting patients. Features may be discrete or continuous, and some features may be missing as not all tests are necessarily performed on all patients. The clinical predictive tasks consist of utilising the routine clinical features $x_i$ to predict, for a newly presenting patient, (i) the likelihood $\hat{y}_\text{SARS-CoV-2}$ of receiving a positive SARS-CoV-2 test result, (ii) the likelihood $\hat{y}_\text{admission}$ of requiring hospital admission, and the (iii) likelihood $\hat{y}_\text{ICU}$ of requiring intensive care. In addition, we are given a development dataset consisting of $N$ patients, their corresponding observed routine clinical features $x_i$, SARS-CoV-2 test results $y_\text{SARS-CoV-2} \in \{0,1\}$, hospital admissions $y_\text{admission} \in \{0,1\}$, and ICU admissions $y_\text{ICU} \in \{0,1\}$, where 1 indicates the presence of an outcome. Using this development dataset, our goal is to derive clinical predictive models $\hat{f}_\text{SARS-CoV-2}$, $\hat{f}_\text{admission}$ and $\hat{f}_\text{ICU}$ for the respective before-mentioned tasks in order to inform care and help prioritise scarce healthcare resources.
\begin{align}
\hat{y}_\text{SARS-CoV-2} &= \hat{f}_\text{SARS-CoV-2}(x) \\
\hat{y}_\text{admission} &= \hat{f}_\text{admission}(x) \\
\hat{y}_\text{ICU} &= \hat{f}_\text{ICU}(x)
\end{align}

\subsubsection{Methodology} To derive the clinical predictive models $\hat{f}_\text{SARS-CoV-2}$, $\hat{f}_\text{admission}$ and $\hat{f}_\text{ICU}$ from the given development dataset, we set up a systematic model development, validation, and evaluation pipeline (Fig. \ref{fig:pipeline}). To evaluate the generalisation ability of the developed clinical predictive models and to rule out overfitting to patients in the evaluation cohort, the development data is initially split into independent and stratified training, validation, and test folds without any patient overlap. Concretely, the multistage pipeline consists of (i) preprocessing, (ii) model development, (iii) model selection, and (iv) model evaluation stages. For preprocessing and model development, only the training fold is used, and only the validation and test folds of the development data are used for model selection and model evaluation, respectively. We outline the pipeline stages in detail in the following paragraphs.

\subsubsection{Preprocessing} In the preprocessing stage, we first drop all input features that are missing for more than 99.8\% of all training set patients to ensure we have a minimal amount of data for each feature. This removes a total of 9 features from the original 106 routine clinical, laboratory and demographic features. We then transform all discrete features for each patient into their one-hot encoded representation with one out of $p$ indicator variables set to 1 to indicate the discrete value for this patient, and all others set to 0 with $p$ being the number of unique values for the discrete feature. We defined those features as discrete that have fewer than 6 unique values across all patients in the training fold. For discrete features, missing features were counted as a separate category in the one hot representation. Next, we standardised all continuous features to have zero mean and unit standard deviation across the training fold data. Lastly, we performed multiple imputation by chained equations (MICE) to impute all missing values of every continuous feature from the respective other features in an iterative fashion \cite{white2011multiple}. We additionally added a missing indicator that indicates 1 if the feature was imputed by MICE and 0 if it was originally present in order to preserve missingness information in the data after imputation. After the preprocessing stage, continuous input features are standardised and fully imputed, and discrete input features are one-hot encoded. All preprocessing operations are derived only from the training fold, and na\"ively applied without adjustment to validation and test folds in order to avoid information leakage.

\subsubsection{Model Development} In the model development stage, we train candidate clinical predictive models $\hat{f}_\text{SARS-CoV-2}$, $\hat{f}_\text{admission}$ and $\hat{f}_\text{ICU}$ using supervised learning on the training fold of the preprocessed data. To derive the models from the preprocessed training fold data, we optimise various types of predictive models, and perform a hyperparameter search with $m$ runs for each of them. The model development process yields $m$ candidate models with different hyperparameter choices and predictive performances for each model category.

\subsubsection{Model Selection} In order to select the best model amongst the set of candidate models, we evaluate their predictive performance against the held-out validation fold that had not been used for model development. We choose the top candidate model by ranking all models by their evaluated predictive performance. The model selection stage using the independent validation fold enables us to optimise hyperparameters without utilising test fold data.

\subsubsection{Model Evaluation} In the model evaluation stage, we evaluate the selected best clinical predictive model against the held-out test fold that had not been used neither for training nor model selection in order to estimate the expected generalisation error of the models on previously unseen data. Using this approach, every selected best model from the model selection stage is evaluated exactly once against the test fold.

Using the presented standardised model development, selection and evaluation pipeline, we compare various types of clinical predictive models in the same test setting, with exactly the same amount of hyperparameter optimisation and input features against the same test fold. This process enables us to systematically study the expected generalisation ability, predictive performance and influential features of clinical predictive models for predicting SARS-CoV-2 test results, hospital admission for SARS-CoV-2 positive patients, and ICU admission for SARS-CoV-2 positive patients.

\section{Experiments}
We conducted retrospective experiments to evaluate the predictive performance of a number of clinical predictive models on each of the presented clinical prediction tasks using the standardised development, validation and evaluation pipeline. Concretely, our experiments aimed to answer the following questions:
\begin{enumerate}
\item[1] What is the expected predictive performance of the various clinical predictive models in predicting (i) SARS-CoV-2 test results for presenting patients, (ii) hospital admission for SARS-CoV-2 positive patients, and (iii) ICU admission for SARS-CoV-2 positive patients?
\item[2] Which clinical, demographic and blood analysis features were most important for the respective best encountered predictive models for each clinical prediction task?
\end{enumerate}

\noindent The following subsections describe the conducted experimental evaluation in detail.
 \begin{table}[t!]
\caption{Population Statistics. \textup{Training, validation, and test fold statistics for all patients (top) and SARS-CoV-2 positive patients (bottom). Age is specified in 20-quantiles (10\% and 90\% quantiles in parentheses).}}
	\label{tb:dataset}
	\centering
	\begin{small}
		\begin{tabular}{l@{\hskip 0.65ex}l@{\hskip 0.65ex}l@{\hskip 0.65ex}l@{\hskip 0.65ex}r}
			\toprule
			\multicolumn{4} {c} {All patients} \\
			Property & Training & Validation & Test \\
			\midrule
			Subjects (\#) & 2822 (50\%) & 1129 (20\%) & 1693 (30\%)\\
			SARS-CoV-2 (\%) & 9.85 & 9.92 & 9.92\\
			Admit (\%) & 1.42 & 1.33 & 1.42\\
			ICU (\%) & 1.59 & 1.68 & 1.59 \\
			Age (quantile) & 9.0 (1.0, 17.0) & 9.0 (1.0, 18.0) & 9.0 (2.0, 17.0)\\ 
			\bottomrule \\
			\toprule
			\multicolumn{4} {c} {SARS-CoV-2 positive patients} \\
			Property & Training & Validation & Test \\
			\midrule
			Subjects (\#) & 279 (50\%) & 112 (20\%) & 167 (30\%)\\
			SARS-CoV-2 (\%) & 100. & 100. & 100.\\
			Admit (\%) & 6.45 & 6.25 & 6.59\\
			ICU (\%) & 2.87 & 2.68 & 2.99 \\
			Age (quantile) & 10.0 (4.0, 17.0) & 11.5 (4.5, 18.5) & 10.0 (4.0, 17.5)\\
			\bottomrule
		\end{tabular}
	\end{small}
\end{table}
\subsection{Dataset and Study Cohort} We used anonymised data from a cohort of 5644 patients seen at the Hospital Israelita Albert Einstein in S\~ao Paulo, Brazil in the early months of 2020\footnote{Exact data collection dates are unknown. The dataset is available at https://www.kaggle.com/einsteindata4u/covid19}. Over the data collection time frame, the rate of SARS-CoV-2 positive patients at the hospital was around 10\% of which around 6.5\% and 2.5\% required hospitalisation and critical care, respectively (Table \ref{tb:dataset}). Notably, younger patients were underrepresented in the SARS-CoV-2 positive group relative to the general patient population which may have been caused by the reportedly more severe disease progression in older patients \cite{remuzzi2020covid}. Information on patient sex was not included in our dataset. We randomly split the entire available patient cohort into training (50\%), validation (20\%) and test folds (30\%) within strata of patient age, SARS-CoV-2 test result, hospital admission status, and ICU admission status. After stratification, the three folds were approximately balanced across the stratification dimensions.

\subsection{Models} Using the presented systematic evaluation methodology, we trained five different model types: Logistic Regression (LR), Neural Network (NN), Random Forest (RF), Support Vector Machine (SVM), and Gradient Boosting (XGB) \cite{chen2016xgboost}. The NN was a multi-layer perceptron (MLP) consisting of $L$ hidden layers with $N$ hidden units each followed by a non-linear activation function (ReLU \cite{nair2010rectified}, SELU \cite{klambauer2017self}, or ELU \cite{clevert2015fast}) and batch normalisation \cite{ioffe2015batch}, and was trained using the Adam optimiser \cite{kingma2014adam} for up to 300 epochs with an early stopping patience of 12 epochs on the validation set loss.

\subsection{Hyperparameters} We followed an unbiased, systematic approach to hyperparameter selection and optimisation. For each type of clinical predictive model, we performed a maximum of 30 hyperparameter optimisation runs with hyperparameters chosen from predefined ranges (Table \ref{tb:hyperparameters}). The performance of each hyperparameter optimisation run was evaluated against the validation cohort. After computing the validation set performance, we selected the best candidate predictive model across the 30 hyperparameter optimisation runs by area under the receiver operator curve for further evaluation against the test set.
\begin{table}[t!] 
\caption{Hyperparameters. \textup{Hyperparameter ranges used for hyperparameter optimisation of Logistic Regression (LR), Neural Network (NN), Random Forest (RF), Support Vector Machine (SVM), and Gradient Boosting (XGB) models for all tasks. Parentheses indicate continuous ranges within the indicated limits sampled uniformly. Comma-delimited lists indicate discrete choices with equal selection probability.}}
	\label{tb:hyperparameters}
	\centering
	\begin{small}
		\begin{tabular}{l@{\hskip 1.25ex}l@{\hskip 0.5ex}r}
			\toprule
			& Hyperparameter & Range / Choices\\
			\midrule
			\parbox[t]{2mm}{\multirow{1}{*}{\rotatebox[origin=c]{90}{LR}}}&Regularization strength $C$ & 0.01, 0.1, 1.0, 10.0\\
			\midrule
			\parbox[t]{2mm}{\multirow{7}{*}{\rotatebox[origin=c]{90}{NN}}}& Number of hidden units $N$ & 16, 32, 64, 128\\
			& Number of layers $L$ & 1, 2, 3\\
			& Activation $a$ & ReLU \cite{nair2010rectified}, SELU \cite{klambauer2017self}, ELU \cite{clevert2015fast}\\
			& Batch size $B$ & 16, 32, 64, 128\\
			& L$2$ regularisation $\lambda_2$ & 0.0, 0.00001, 0.0001\\
			& Learning rate $\alpha$ & 0.003, 0.03\\
			& Dropout percentage $p$ & (0\%, 25\%)\\
			\midrule
			\parbox[t]{2mm}{\multirow{2}{*}{\rotatebox[origin=c]{90}{RF}}}&Tree depth $D$ & 3, 4, 5\\
			& Number of trees $T$ & 32, 64, 128, 256\\
			\midrule
			\parbox[t]{2mm}{\multirow{3}{*}{\rotatebox[origin=c]{90}{SVM}}}&Regularization strength $C$ & 0.01, 0.1, 1.0, 10.0\\
			& Kernel $k$ & polynomial, RBF, sigmoid\\
			& Polynomial degree $d$ & 3, 5, 7\\
			\midrule
			\parbox[t]{2mm}{\multirow{7}{*}{\rotatebox[origin=c]{90}{XGB}}}&Subsample ratio $r$ & 0.25, 0.5, 0.75, 1.0\\
			& Max. tree depth $T$ & 2, 3, 4, 5, 6, 7, 8\\
			& Min. partition loss $\gamma$ & 0.0, 0.1, 1.0, 10.0\\
			& Learning rate $\alpha$ & 0.003, 0.03, 0.3, 0.5\\
			& L1 regularisation $\lambda_1$ & 1.0, 0.1, 0.001, 0.0\\
			& L2 regularisation $\lambda_2$ & 1.0, 0.1, 0.001, 0.0\\
			& Num. boosting rounds $B$ & 5, 10, 15, 20\\
			\bottomrule
		\end{tabular}
	\end{small}
\end{table}

\begin{table*}[t!] 
\caption{Predictive Performance. \textup{Comparison of Logistic Regression (LR), Neural Network (NN), Random Forest (RF), Support Vector Machine (SVM), and Gradient Boosting (XGB) models in terms of AUC, AUPR, sensitivity, specificity, and specificity at greater than 95\% sensitivity (Spec.@95\%Sens.) for predicting (i) SARS-CoV-2 test results, (ii) hospital admission for SARS-CoV-2 positive patients, and (iii) ICU admission for SARS-CoV-2 positive patients on the test set cohort. Best results in bold. In parentheses are the 95\% confidence intervals (CIs) obtained via bootstrap resampling with 100 samples. \sig = significant at $\alpha = 0.05$ (t-test) to the model with the highest predictive performance in terms of AUC.}}
	\label{tb:results_all}
	\centering
	\begin{small}
		\begin{tabular}{l@{\hskip 17.0ex}l@{\hskip 5ex}l@{\hskip 5ex}l@{\hskip 5ex}l@{\hskip 5ex}l}
			\toprule
			\multicolumn{6} {c} {\raisebox{-.15\height}{\includegraphics[height=10pt]{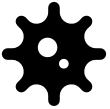}} (i) SARS-CoV-2 Test Results} \\ \\
			Model & AUC & AUPR & Sensitivity & Specificity & Spec.@95\% Sens.\\
			\midrule
XGB & \hspace{1ex}\textbf{0.66} (0.63, 0.70) & \hspace{1ex}{0.21} (0.15, 0.28) & \hspace{1ex}\textbf{0.75} (0.67, 0.81) & \hspace{1ex}{0.49} (0.46, 0.51) & \hspace{1ex}\textbf{0.23} (0.07, 0.32) \\
RF & \sig{}{0.65} (0.62, 0.69) & \sig{}{0.19} (0.14, 0.24) & \sig{}{0.69} (0.61, 0.74) & \sig{}{0.54} (0.51, 0.57) & \sig{}{0.19} (0.10, 0.25) \\
NN & \sig{}{0.62} (0.57, 0.65) & \sig{}\textbf{0.22} (0.15, 0.28) & \sig{}{0.60} (0.52, 0.67) & \sig{}{0.55} (0.53, 0.58) & \sig{}{0.17} (0.14, 0.28) \\
LR & \sig{}{0.61} (0.57, 0.65) & \sig{}{0.17} (0.13, 0.24) & \sig{}{0.58} (0.51, 0.65) & \sig{}{0.55} (0.52, 0.57) & \sig{}{0.19} (0.16, 0.25) \\
SVM & \sig{}{0.61} (0.57, 0.65) & \hspace{1ex}{0.21} (0.15, 0.27) & \sig{}{0.57} (0.51, 0.64) & \sig{}\textbf{0.59} (0.56, 0.61) & \sig{}{0.14} (0.06, 0.16) \\
			\bottomrule \\
			\toprule
			\multicolumn{6} {c} {\raisebox{-.15\height}{\includegraphics[height=10pt]{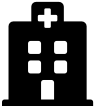}} (ii) Hospital Admission for SARS-CoV-2 Positive Patients} \\ \\
			Model & AUC & AUPR & Sensitivity & Specificity & Spec.@95\% Sens.\\
			\midrule
RF & \hspace{1ex}\textbf{0.92} (0.81, 0.98) & \hspace{1ex}{0.43} (0.19, 0.81) & \hspace{1ex}{0.55} (0.19, 0.85) & \hspace{1ex}\textbf{0.96} (0.92, 0.98) & \hspace{1ex}\textbf{0.34} (0.29, 0.97) \\
XGB & \hspace{1ex}{0.91} (0.80, 0.98) & \sig{}\textbf{0.52} (0.28, 0.84) & \sig{}{0.64} (0.43, 0.95) & \sig{}{0.94} (0.90, 0.97) & \sig{}{0.00} (0.00, 0.94) \\
LR & \sig{}{0.88} (0.70, 0.98) & \hspace{1ex}{0.44} (0.18, 0.83) & \sig{}\textbf{0.82} (0.52, 1.00) & \sig{}{0.85} (0.79, 0.90) & \sig{}{0.13} (0.08, 0.93) \\
NN & \sig{}{0.85} (0.68, 0.97) & \sig{}{0.31} (0.13, 0.66) & \sig{}{0.64} (0.33, 1.00) & \sig{}{0.95} (0.91, 0.97) & \sig{}{0.11} (0.06, 0.93) \\
SVM & \sig{}{0.85} (0.70, 0.98) & \sig{}{0.35} (0.17, 0.77) & \sig{}{0.64} (0.30, 1.00) & \sig{}{0.95} (0.91, 0.97) & \sig{}{0.21} (0.15, 0.96) \\
			\bottomrule \\
			\toprule
			\multicolumn{6} {c} {\raisebox{-.15\height}{\includegraphics[height=10pt]{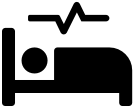}} (iii) Critical Care Admission for SARS-CoV-2 Positive Patients} \\ \\
			Model & AUC & AUPR & Sensitivity & Specificity & Spec.@95\% Sens.\\
			\midrule
SVM & \hspace{1ex}\textbf{0.98} (0.95, 1.00) & \hspace{1ex}{0.53} (0.14, 1.00) & \hspace{1ex}\textbf{0.80} (0.36, 1.00) & \hspace{1ex}{0.96} (0.92, 0.98) & \hspace{1ex}\textbf{0.95} (0.91, 1.00) \\
LR & \hspace{1ex}\textbf{0.98} (0.93, 1.00) & \sig{}\textbf{0.67} (0.09, 1.00) & \hspace{1ex}\textbf{0.80} (0.29, 1.00) & \sig{}{0.93} (0.89, 0.96) & \sig{}{0.91} (0.87, 1.00) \\
NN & \sig{}{0.97} (0.94, 0.99) & \sig{}{0.35} (0.10, 0.88) & \hspace{1ex}\textbf{0.80} (0.36, 1.00) & \sig{}{0.95} (0.91, 0.99) & \hspace{1ex}{0.94} (0.90, 0.99) \\
RF & \hspace{1ex}{0.97} (0.92, 1.00) & \sig{}{0.56} (0.13, 1.00) & \sig{}{0.60} (0.15, 1.00) & \sig{}\textbf{0.98} (0.96, 1.00) & \sig{}{0.90} (0.86, 1.00) \\
XGB & \sig{}{0.67} (0.53, 0.98) & \sig{}{0.29} (0.01, 0.68) & \sig{}{0.40} (0.00, 1.00) & \sig{}{0.94} (0.91, 0.97) & \sig{}{0.00} (0.00, 0.96) \\
			\bottomrule
		\end{tabular}
	\end{small}
\end{table*}

\subsection{Metrics}
\subsubsection{Predictive Performance} 
To assess the predictive performance of each of the developed clinical predictive models, we evaluated their performance in terms of area under the receiver operator curve (AUC),  area under the precision recall curve (AUPR), sensitivity, specificity, and specificity at greater than 95\% sensitivity (Spec.@95\%Sens.) on the held-out test set cohorts for each task (Table \ref{tb:dataset}). After model development and hyperparameter optimisation, we evaluated each model type exactly once against the test set to calculate the final performance metrics. Operating thresholds for each model were the operating points on the receiver operator characteristic curve closest to the top left coordinate as calculated for the validation cohort. We chose a variety of complementary evaluation metrics in order to give a comprehensive picture of the expected performance of each clinical predictive model on the evaluated tasks. For each of the performance metrics, we additionally computed 95\% confidence intervals (CIs) using bootstrap resampling with 100 bootstrap samples on the test set cohort in order to quantify the uncertainty of our analysis results. We also assessed whether differences between clinical predictive models were statistically significant at significance level $\alpha = 0.05$ using pairwise t-tests with the respective best models for each task as measured by AUC.

\subsubsection{Importance of Test Types}
To quantify the importance of specific clinical, demographic and blood analysis features on each of the predicted outcomes, we utilised causal explanation (CXPlain) models \cite{schwab2019cxplain}. CXPlain provides standardised relative feature importance attributions for any predictive model by computing the marginal contribution of each input feature towards the predictive performance of a model \cite{schwab2018granger}, and is therefore particularly well-suited for assessing feature importance in our diverse set of models. We used the test fold's ground truth labels to compute the exact marginal contribution of each input feature without any estimation uncertainty.

\section{Results}
\subsection{Predictive Performance} 
In terms of predictive performance (Table \ref{tb:results_all}), we found that the overall best identified models by AUC were XGB for predicting SARS-CoV-2 test results, RF for predicting hospital admissions for SARS-CoV-2 positive patients, and SVM for predicting ICU admission for SARS-CoV-2 positive patients with respective AUCs of 0.66 (95\% CI: 0.63, 0.70), 0.92 (95\% CI: 0.81, 0.98), and 0.98 (95\% CI: 0.95, 1.00). Notably, we found that predicting positive SARS-CoV-2 results from routinely collected clinical measurements was a considerably more difficult task for clinical predictive models than predicting hospitalisation and ICU admission. Nonetheless, the best encountered clinical predictive model for predicting SARS-CoV-2 test results (XGB) achieved a respectable sensitivity of 75\% (95\% CI: 67\%, 81\%) and specificity of 49\% (95\% CI: 46\%, 51\%). After fixing the operating threshold of the model to meet a sensitivity level of at least 95\% (Spec.@95\% Sens.), the best XGB model for predicting SARS-CoV-2 test results would achieve a specificity of 23\% (95\% CI: 7\%, 32\%). We additionally found that the differences in predictive performance between the best XGB model for predicting SARS-CoV-2 test results and the other predictive models was significant at a pre-specified significance level of $\alpha = 0.05$ (t-test) for all but the AUPR metric, where NN achieved a significantly better AUPR of 0.22 and the difference to SVM was not significant at the pre-specified significance level. On the task of predicting hospital admissions for SARS-CoV-2 positive patients, the best encountered RF model achieved a sensitivity of 55\% (95\% CI: 19\%, 85\%), a high specificity of 96\% (95\% CI: 92\%, 98\%), and a specificity at a fixed sensitivity of at least 95\% (Spec.@95\% Sens.) of 34\% (95\% CI: 29\%, 97\%). Owing to the lower sample size due to the smaller cohort of SARS-CoV-2 positive patients, the performance results for predicting hospital admission generally had wider uncertainty bounds but were nonetheless significantly better for RF than the other predictive models at the pre-specified significance level of $\alpha = 0.05$ (t-test) for most performance metrics with the exception of AUC where XGB achieved an AUC of 0.91 and AUPR where LR achieved an AUPR of 0.44. On the task of predicting ICU admission for SARS-CoV-2 positive patients, SVM had a sensitivity of 80\% (95\% CI: 36\%, 100\%), a specificity of 96\% (95\% CI: 92\%, 98\%), and a specificity at a fixed sensitivity of at least 95\% (Spec.@95\% Sens.) of 95\% (95\% CI: 91\%, 100\%). Due to the small percentage of around 3\% of SARS-CoV-2 positive patients that were admitted to the ICU (Table \ref{tb:dataset}), uncertainty bounds were wider than for the models predicting hospital admissions, and the results of the best encountered SVM were found to be not significantly better than LR and RF in terms of AUC, LR and NN in terms of sensitivity, and NN in terms of Spec.@95\% Sens. at the pre-specified significance level of $\alpha = 0.05$ (t-test).

\subsection{Feature Importance} In terms of feature importance, we found that importance scores were distributed highly unequally, relatively uniform and highly uniform for the best models encountered for predicting SARS-CoV-2 test results, for predicting hospital admissions for SARS-CoV-2 positive patients, and for predicting ICU admission, respectively (Figure \ref{fig:importance}). Most notably, we found that 71.7\% of the importance for the best XGB model for predicting SARS-CoV-2 test results was assigned to the missing indicator corresponding to the Arterial Lactic Acid measurement, i.e. much of the marginal predictive performance gain of the XGB model was attributed to whether or not the Arterial Lactic Acid test had been ordered. Beyond Arterial Lactic Acid being missing, age, leukocyte count, platelet count, and creatinine were implied to be associated with a positive SARS-CoV-2 test result by the best encountered predictive model, which further substantiates recent independent reports of those factors being potentially associated with SARS-CoV-2 \cite{covid2020severe,rothan2020epidemiology,lippi2020thrombocytopenia,cheng2020kidney,wang2020clinical}. Similarly to the best encountered XGB model for predicting SARS-CoV-2 test results, the top encountered predictive models for hospital admission and ICU admission for SARS-CoV-2 positive patients assigned a considerable degree of importance to missingness patterns associated with a number of measurements. A possible explanation for missingness appearing as a top predictor across the different tasks is that decisions whether or not to order a certain test to be performed for a given patient were influenced by patient characteristics that were not captured in the set of clinical measurements that were available to the predictive models. A controlled setting with standardised testing guidelines would be required to determine which confounding factors are behind the predictive power of the missingness patterns that have been implied to be associated with COVID-19 by the predictive models. Beyond missingness patterns, top predictors for predicting hospital admission were lactate dehydrogenase \cite{zhou2020clinical}, gamma-glutamyltransferase, which through abnormal liver function has been reported to be implicated in COVID-19 severity \cite{fan2020clinical}, and HCO$_3$ \cite{dondorp2020respiratory}. For predicting ICU admission in SARS-CoV-2 positive patients, pCO$_2$ and pH \cite{wang2020clinical} were top predictors. Blood pH, and in particular respiratory alkalosis, has been reported to be associated with severe COVID-19 \cite{ashraf2020covid}.

\section{Discussion}
We presented a systematic study of predictive models that predict SARS-CoV-2 test results, hospital admission for SARS-CoV-2 positive patients, and ICU admission for SARS-CoV-2 positive patients using routinely collected clinical measurements. Models that predict SARS-CoV-2 test results could help prioritise scarce testing capacity by identifying those individuals that are more likely to receive a positive result. Similarly,  predictive models that predict which SARS-CoV-2 positive patients would be most likely to require hospital and critical care beds could help better utilise existing hospital capacity by prioritising those patients that have the highest risk of deterioration. Facilitating the efficient utilisation of scarce healthcare resources is particularly important in dealing with SARS-CoV-2 as its rapid transmission significantly increases demand for healthcare services worldwide.

\begin{figure}[t!]
  \centering
      \includegraphics[width=0.91\linewidth]{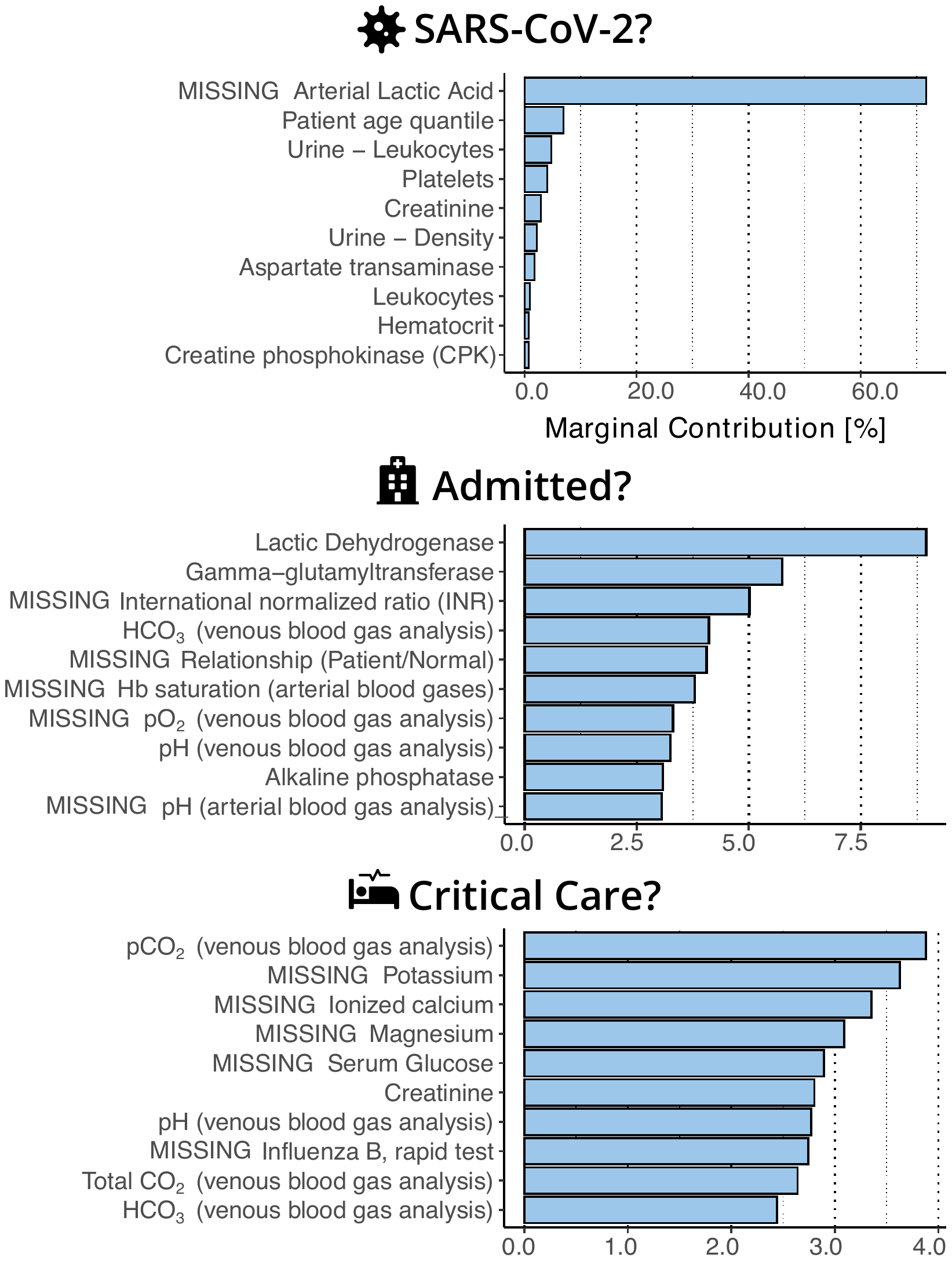}
      \caption{A comparison of the top-10 features ranked by relative feature importance scores for the best encountered model for predicting SARS-CoV-2 test results (XGB, top), hospital admission for SARS-CoV-2 positive patients (RF, middle), and critical care admission for SARS-CoV-2 positive patients (SVM, bottom), respectively. The bar length corresponds to the relative marginal importance (in \%) of the displayed features towards the predictive performance of the respective model. Feature names that include an all-caps "MISSING" indicate that the given marginal contribution refers to the importance of the absence or presence of that feature, not the feature itself.} 
\label{fig:importance}
\end{figure}

The main limitation of the presented study is that its experimental evaluation was based on data collected from a single study site, and its results may therefore not generalise to settings with significantly different patient populations, admission criteria, patterns of missingness, and testing guidelines. In addition, we did not have access to mortality data for the analysed cohort, and we were therefore not able to correlate our predicted individual risk scores with patient mortality, which is another related prediction task that may be of clinical importance. Future studies should include a broader set of clinical measurements and outcomes, cohorts from multiple distinct geographical sites and under varying patterns of missingness in order to determine the robustness of the clinical predictive models to these confounding factors. Finally, we believe that the inclusion of data from other modalities, such as genomic profiling and medical imaging, and data on co-morbidities, symptoms and treatment histories could potentially further improve predictive performance of clinical predictive models across the presented prediction tasks.

\section{Conclusion}
We presented a systematic study in which we developed and evaluated clinical predictive models for COVID-19 that estimate (i) the likelihood of a positive SARS-CoV-2 test in patients presenting at hospitals, (ii) the likelihood of hospital admission and (iii) intensive care unit admission in SARS-CoV-2 positive patients. We evaluated our developed clinical predictive models in a retrospective evaluation using a cohort of 5644 hospital patients seen in S\~ao Paulo, Brazil. In addition, we determined the clinical, demographic and blood analysis measurements that were most important for accurately predicting SARS-CoV-2 status, hospital admissions, and ICU admissions. Our experimental results indicate that clinical predictive models may in the future potentially be used to inform care and help prioritise scarce healthcare resources by assigning personalised risk scores for individual patients using routinely collected clinical, demographic and blood analysis data. Furthermore, our findings on the importance of routine clinical measurements towards predicting clinical pathways for patients increase our understanding of the interrelations of individual risk profiles and outcomes in SARS-CoV-2. Based on our study's results, we conclude that healthcare systems should explore the use of predictive models that assess individual COVID-19 risk in order to improve healthcare resource prioritisation and inform patient care. 

\begin{figure}[t!]
  \centering
      \includegraphics[width=0.95\linewidth]{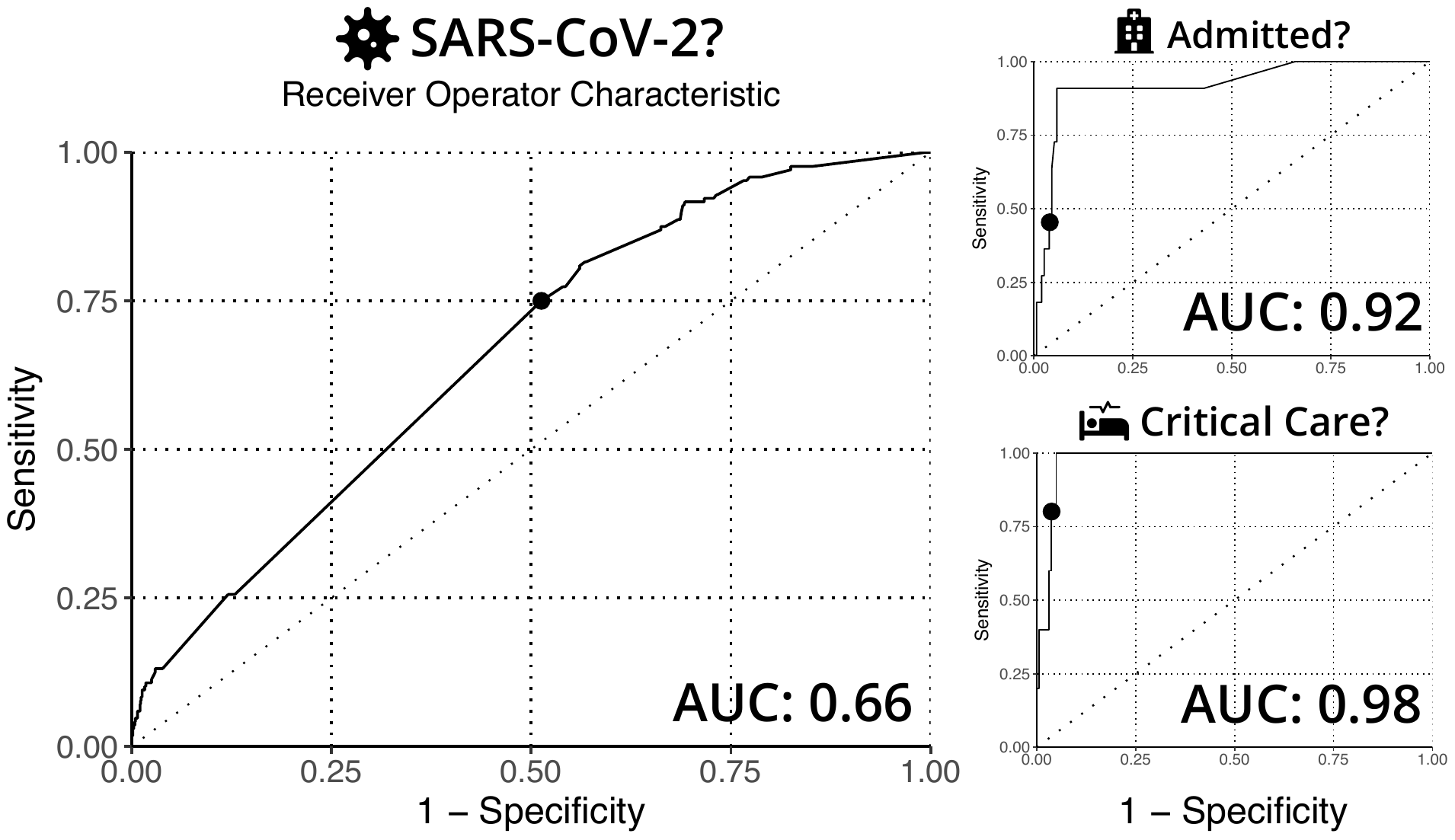}
      \caption{Receiver operator characteristic (ROC) curves for the best encountered model for predicting SARS-CoV-2 test results (XGB, left), hospital admissions for SARS-CoV-2 positive patients (RF, top right), and critical care admissions for SARS-CoV-2 positive patients (SVM, bottom right). Numbers in the bottom right of each subgraph show the respective model's area under the curve (AUC). Solid dots on the curves indicate operating thresholds selected on the validation fold.} 
\label{fig:importance}
\end{figure}

\appendices

\section*{Acknowledgments}
The anonymised data used in this manuscript were generously contributed by patients at Hospital Israelita Albert Einstein in S\~ao Paulo, Brazil, and are freely available at https://www.kaggle.com/einsteindata4u/covid19.

\small
\bibliographystyle{IEEEtran}
\bibliography{references.bib}

\end{document}